\documentclass[journal]{IEEEtran}
\usepackage{cite}
\usepackage[pdftex]{graphicx}

\usepackage{amsmath}
\usepackage{url}

\usepackage{calc}
\newlength{\arrayrulewidthOriginal}

\usepackage[hidelinks=true]{hyperref}
\usepackage{multirow}
\usepackage{amsmath}
\usepackage{amsthm}
\usepackage{amssymb}
\usepackage{amsbsy}
\usepackage{cleveref}
\usepackage{algorithm}
\usepackage{algpseudocode}
\usepackage[caption=false, font=footnotesize]{subfig}

\usepackage[table]{xcolor}
\usepackage{graphicx}
\usepackage{upgreek}
\usepackage{array}
\usepackage{booktabs}
\usepackage{makecell}
\usepackage{tabularx}
\usepackage{adjustbox}
\usepackage{xcolor}
\usepackage{cite}
\usepackage{pifont}
\usepackage[normalem]{ulem}
\usepackage{fp}
\usepackage[table]{xcolor}
\usepackage{tikz}
\usepackage{color, soul}
\usepackage{xcolor}

\usepackage{algorithm}
\usepackage{algpseudocode}
\usepackage{textpos}
\usepackage{multirow}
\usepackage{tikz}
\usepackage{units}
\usepackage{setspace}
\usepackage{siunitx}
\usepackage{epstopdf}
\usepackage{cleveref}
\usepackage{hyperref}

\usetikzlibrary{calc,shadings,patterns,tikzmark}

\newcommand\percentage[2][round-precision = 2]{
    \SI[round-mode = places,
        scientific-notation = fixed, fixed-exponent = 0,
        output-decimal-marker={.}, #1]{#2e2}{}
}

\newcommand\percentageBest[2][round-precision = 2]{
    {\bfseries \SI[round-mode = places,
        scientific-notation = fixed, fixed-exponent = 0, detect-inline-weight  = math, detect-weight = true,
        output-decimal-marker={.}, #1]{#2e2}{}}
}

\newcommand\round[2][round-precision = 3]{
    \SI[round-mode = places, #1]{#2}{}
}

\newcommand\roundBest[2][round-precision = 3]{
    {\bfseries \SI[round-mode = places, detect-inline-weight  = math, detect-weight = true, #1]{#2}{}}
}

\NewDocumentCommand{\rot}{O{45} O{1em} m}{\hspace{-1cm} \makebox[#2][l]{\rotatebox{#1}{#3}}}%

\definecolor{ben}{HTML}{326B34}

\begin{document}
\title{BigEarthNet-MM: A Large Scale Multi-Modal Multi-Label Benchmark Archive for Remote Sensing Image Classification and Retrieval}

\author{Gencer~Sumbul,~\IEEEmembership{Graduate Student Member,~IEEE,}
Arne~de~Wall,~\IEEEmembership{Student Member,~IEEE,}
        Tristan~Kreuziger,~\IEEEmembership{Student Member,~IEEE,}
        Filipe~Marcelino,
        Hugo~Costa,
        Pedro~Benevides,
        M{\'a}rio~Caetano,
        Beg\"um~Demir,~\IEEEmembership{Senior Member,~IEEE,}
        Volker~Markl
\thanks{Gencer~Sumbul, Arne~de~Wall, Tristan~Kreuziger, Beg\"um~Demir and Volker Markl are with Technische Universit\"at Berlin, Berlin, Germany.}
\thanks{Filipe~Marcelino, Hugo~Costa, Pedro~Benevides, and M{\'a}rio~Caetano are with Direção-Geral do Território (DGT), Lisbon, Portugal. Hugo~Costa and M{\'a}rio~Caetano are also with NOVA Information Management School (NOVA IMS), Universidade Nova Lisboa, Campus de Campolide, 1070-312 Lisbon, Portugal.}}

\maketitle
\begin{abstract}
This paper presents the multi-modal BigEarthNet (BigEarthNet-MM) benchmark archive made up of 590,326 pairs of Sentinel-1 and Sentinel-2 image patches to support the deep learning (DL) studies in multi-modal multi-label remote sensing (RS) image retrieval and classification. Each pair of patches in BigEarthNet-MM is annotated with multi-labels provided by the CORINE Land Cover (CLC) map of 2018 based on its thematically most detailed Level-3 class nomenclature. Our initial research demonstrates that some CLC classes are challenging to be accurately described by only considering (single-date) BigEarthNet-MM images. In this paper, we also introduce an alternative class-nomenclature as an evolution of the original CLC labels to address this problem. This is achieved by interpreting and arranging the CLC Level-3 nomenclature based on the properties of BigEarthNet-MM images in a new nomenclature of 19 classes. In our experiments, we show the potential of BigEarthNet-MM for multi-modal multi-label image retrieval and classification problems by considering several state-of-the-art DL models. We also demonstrate that the DL models trained from scratch on BigEarthNet-MM outperform those pre-trained on ImageNet, especially in relation to some complex classes, including agriculture and other vegetated and natural environments. We make all the data and the DL models publicly available at \url{https://bigearth.net}, offering an important resource to support studies on multi-modal image scene classification and retrieval problems in RS.
\end{abstract}


\IEEEpeerreviewmaketitle

\section{Introduction}
As a result of advancements in satellite technology, recent years have witnessed a significant increase in the volume of remote sensing (RS) image archives. Accordingly, the development of accurate scene classification and content based image retrieval (CBIR) systems in massive image archives has attracted great attention in RS. CBIR systems aim to
achieve an efficient and precise retrieval of RS images from large archives that are similar to a query image~\cite{Roy:2020}, \cite{8067633}. RS image scene classification systems aim at automatically assigning class labels to each RS image scene in a large archive~\cite{Sumbul:2019}, \cite{9127795}. Deep learning (DL) based methods have recently seen a rise in popularity in the context of RS image scene classification and retrieval problems. Most DL models require a high amount of annotated images during training to optimize all parameters and reach a high performance. The availability and quality of such data determine the feasibility of many DL models. There are several benchmark archives made publicly available for different RS applications (e.g., pixel-based image classification). For a comprehensive list, we refer the reader to \cite{sen12-paper}. To the best of our knowledge, most of the existing publicly available benchmark archives for image scene classification and retrieval problems contain: 1) single-modal RS images (e.g., multispectral or SAR); and 2) single-label image annotations (i.e., each image is annotated by a single label that is associated with the most significant content of the considered image). However, multi-modal images associated with the same geographical area allow for rich characterization of RS images and thus improve image retrieval performance when jointly considered~\cite{Hong:2020}. In addition, RS images usually contain areas with a high variety of semantically complex content that must be reflected by more than one class annotation through multiple class labels (multi-labels). 

Thus, a benchmark archive consisting of multi-modal images annotated with multi-labels is needed. However, annotating RS images with multi-labels at a large-scale to drive DL studies is time consuming, complex, and costly in operational scenarios. To overcome this problem, a common approach is to exploit DL models with proven architectures (such as ResNet \cite{he2016deep} or VGG \cite{simonyan2014very}), which are pre-trained on publicly available general purpose datasets in the computer vision (CV) community (e.g., ImageNet \cite{imagenet}). The existing model is then fine-tuned on a small set of RS images annotated with multi-labels to calibrate the final layers. This strategy is also known as a transfer learning strategy. There are several versions of the above-mentioned models that have been pre-trained on large-scale datasets in CV. However, we argue that this is not a proper approach in RS, because of the differences in image characteristics in CV and RS. For example, Sentinel-2 multispectral images have 13 spectral bands associated with varying and lower spatial resolutions compared to the CV images. In addition, the semantic content present in CV and RS images is significantly different, and thus the respective semantic classes differ from each other. To address this issue, we have recently introduced BigEarthNet \cite{bigearthnet} as a large-scale single-modal benchmark archive for RS image search, retrieval and classification. BigEarthNet contains 590,326 Sentinel-2 image patches annotated with multi-labels provided by the CORINE Land Cover (CLC) map of 2018 (CLC 2018) \cite{corine}. The CLC nomenclature includes land cover and land use classes grouped in a three-level hierarchy, and for the BigEarthNet image patches, the most thematically detailed Level-3 class nomenclature is considered. However, there are some CLC classes that are difficult to be identified by only exploiting (single-date) Sentinel-2 images, because: i) land use concepts associated with some classes (e.g., \textit{Dump sites}, \textit{Sport and leisure facilities}) may not be visible from space or fully recognizable with the spatial resolution of Sentinel-2 images, and ii) RS time series, which BigEarthNet does not include, may be required to describe and discriminate some classes (e.g., \textit{Non-irrigated arable land}, \textit{Permanently irrigated land}). In addition, BigEarthNet is not suitable for the multi-modal learning-based algorithm development and validation purposes, since it only contains Sentinel-2 image patches. 

To overcome these issues, in this paper we introduce the multi-modal BigEarthNet (BigEarthNet-MM) that contains 590,326 pairs of Sentinel-2 and Sentinel-1 image patches. We also introduce an alternative nomenclature for images in BigEarthNet-MM as an evolution of the original CLC labels. Fig.~\ref{fig:retrieval} shows an example of the BigEarthNet-MM image pairs and their multi-labels from the new nomenclature.

\section{Description of BigEarthNet-MM}
 BigEarthNet-MM contains 590,326 pairs of Sentinel-1 and Sentinel-2 image patches acquired over $10$ different European countries (Austria, Belgium, Finland, Ireland, Kosovo, Lithuania, Luxembourg, Portugal, Serbia, Switzerland). Sentinel-2 patches of BigEarthNet-MM are taken from the original BigEarthNet \cite{bigearthnet}. To construct these patches, $125$ Sentinel-2 tiles associated with less than 1\% of cloud cover and acquired between June 2017 and May 2018 were considered. All tiles were atmospherically corrected by employing Sentinel-2's Level 2A product generation and formatting tool (sen2cor) provided by the European Space Agency due to its proven success in the literature. After the atmospheric correction, the 10\textsuperscript{th} band of each image patch is not available anymore, as it is the cirrus band (which is omitted in the Level 2A output for its lack of surface information). Then, the tiles were divided into $590,326$ non-overlapping image patches, each of which is a section of: 1) $120\times120$ pixels for 10m bands; 2) $60\times60$ pixels for 20m bands; and 3) $20\times20$ pixels for 60m bands. One important goal during the tile selection process was to represent all chosen geographic locations with images acquired in different seasons. Due to the restrictions of finding tiles with a low cloud cover percentage in the relatively narrow time period, this has not been possible at each considered location. Accordingly, the following respective numbers of patches for autumn, winter, spring, and summer have been considered: $143557$, $72877$, $175937$, and $126913$. For the quality check of patches, visual inspection was also employed, which led to the identification of $70,987$ Sentinel-2 image patches that are fully covered by seasonal snow, cloud, and cloud shadow\footnote{The lists are available at \url{http://bigearth.net/\#downloads}.}. 

To construct the Sentinel-1 patches of BigEarthNet-MM, 325 Sentinel-1 Ground Range Detected (GRD) products acquired between June 2017 and May 2018 that jointly cover the area of all original 125 Sentinel-2 tiles with close temporal proximity were selected and processed. The selected scenes provide dual-polarized information channels (VV and VH) and are based on the interferometric wide swath (IW) mode, which is the main acquisition mode over land. All scenes were pre-processed by using the Sentinel-1 toolbox (S1TBX) and the graph processing framework (GPF) of ESA’s Sentinel Application Platform (SNAP). This includes the application of precise orbit files, border and thermal noise removal, radiometric calibration, and geometric correction (i.e., Range Doppler terrain correction). Depending on the spatial extent of the scene, either the SRTM 30 (for scenes below 60° latitude) or the ASTER DEM (for scenes above 60° latitude, where no SRTM 30 exists) were employed in the geometric correction to project images from slant range to ground range. Finally, the backscatter coefficient was converted to a decibel (dB) scale. It is worth noting that, since the selection of the speckle filter is considered to be application dependent, no speckle filtering was applied in our pre-processing workflow in order to preserve the full resolution. This approach is also recommended by the Product Family Specification for SAR of the CEOS Analysis Ready Data for Land (CARD4L) framework\footnote{https://ceos.org/ard/}. 
Based on the pre-processed Sentinel-1 scenes, for each Sentinel-2 patch, a corresponding Sentinel-1 patch with a close timestamp was extracted. In addition, each Sentinel-1 patch inherited the annotations of the corresponding Sentinel-2 patch. The resulting Sentinel-1 image patches contain a spatial resolution of 10m.

\subsection{Class-Nomenclature of BigEartNet-MM}
Each pair (which is made up of Sentinel-1 and Sentinel-2 image patches acquired in the same geographical area) in BigEarthNet-MM is associated with one or more class labels (i.e. multi-labels) extracted from the CORINE land cover map of 2018. CORINE land cover (CLC) is a pioneer adventure initiated in the 80’s of the last century to produce harmonized land cover land use (LCLU) maps in vector format for the member states of the European Union. According to the validation report of the CLC, the accuracy is around $85\%$ \cite{corine-clc-report}. Nowadays, CLC covers 39 countries from Europe and was produced for five reference years, 1990, 2000, 2006, 2012, and 2018. The latter was produced with data of 2017-2018, which matches the time frame of the images included in BigEarthNet. Motivations for embracing a large-scale mapping endeavor aimed at meeting the demand for spatially explicit and harmonized information on land for a variety of purposes, such as environmental management and decision making. The crude state-of-the-art of the 1980’s technology and the large spectrum of potential uses of the maps led to the definition of a coarse spatial resolution and a nomenclature with some broad class definitions, mixing land cover and land use concepts. These definitions are implemented for map production by visual interpretation of RS images and additional data in most countries. Additional data may include very high spatial resolution imagery and official spatial data sets like land registers, often to infer the land use. The same technical specifications were preserved in map updating for historical consistency. Thus the produced five CLC maps have a minimum mapping unit of 25 ha and a minimum mapping width of 100 m, and provide information on land according to a hierarchical nomenclature of 44 classes at the most detailed level (Level3). The image patches in BigEarthNet-MM are representative of 43 CLC classes. In the case that CLC maps are considered as labeling sources for training the machine learning methods to automatically analyse RS images, the modified versions of the CLC nomenclature (which better fit the purpose of the considered application) are commonly preferred. One of the main reason is that RS systems directly observe the land cover rather than the land use. The CLC land-use based labels may not be fully recognizable through the RS images unless they are not associated to very high spatial resolution. As an example, in \cite{paris2019novel} CLC is used as a basis to collect training data for supervised RS image classification, but classes such as \textit{Discontinuous urban fabric} and \textit{Sport and leisure facilities} that depend mainly on land use were removed. A deep revision of the CLC program is actually under consideration following the concept of the EIONET Action Group on Land monitoring in Europe (EAGLE) \cite{arnold2013eagle}. 

To pay more justice to the properties of BigEarthNet-MM image pairs, we introduce a new class-nomenclature by modifying the multi-labels extracted from the CLC 2018. To this end, the CLC Level-3 nomenclature is interpreted and arranged in a new nomenclature of 19 classes\footnote{https://bigearth.eu/BigEarthNetListofClasses.pdf}. Ten classes of the original CLC nomenclature are maintained in the new nomenclature, 22 classes are grouped into 9 new classes, and 11 classes are removed. The classes maintained are semantically homogeneous and largely related to land cover, such as \textit{Broad-leaved forest} and \textit{Beaches, dunes, sands}. Furthermore, CLC classes that are not feasible to be identified by only using single-date BigEarthNet-MM images removed, such as \textit{Burnt areas}. Complex classes (which are often removed when undertaking image classification) are maintained, such as \textit{Complex cultivation patterns} and \textit{Land principally occupied by agriculture, with significant areas of natural vegetation}. The goal is to investigate the ability of DL models to learn from spatial patterns that express semantic classes. Classes are grouped when sharing similar land cover types and spectral patterns. For example, \textit{Moors and heath land} and \textit{Sclerophyllous vegetation} are grouped in a single class, and a new class, \textit{Arable land}, groups similar crops that require dense time series (which not available in BigEarthNet-MM) for their discrimination (e.g. irrigated and non-irrigated crops). Classes that strongly depend on land use or need additional data for their discrimination are removed. For example, class \textit{Airports} essentially relates to land use, and \textit{Intertidal ﬂats} appear in RS images either with or without water depending on the image acquisition time and hence require appropriate time series for its classification. The number of labels associated with each image pair varies between 1 and 12, while $96.80$\% of image pairs are not associated with more than $5$ labels. Only $23$ image pairs are annotated with more than $9$ labels.

\section{Experiments}
\subsection{Experimental Design}
The experiments were carried out in the context of content based multi-modal multi-label RS image retrieval and classification. To achieve multi-modal learning, we stacked the VV and VH bands of Sentinel-1 image patches, and the Sentinel-2 bands associated with $10$m and $20$m spatial resolution into one volume for each pair in BigEarthNet-MM. To this end, we initially applied cubic interpolation to $20$m bands of Sentinel-2 image patches. In the experiments, we did not use the Sentinel-2 image bands associated with $60$m spatial resolution (bands 1 and 9). This is due to the fact that these bands are mainly used for cloud screening, atmospheric correction, and cirrus detection in RS applications and do not embody a significant amount of information for the characterization of semantic content of RS images. In the experiments, we considered the VGG model~\cite{simonyan2014very} and the ResNet model~\cite{he2016deep} at various number of layers (VGG16, VGG19, ResNet50, ResNet101, ResNet152).
To fairly compare all models, we utilized the Adam optimizer~\cite{Adam:2014} with an initial learning rate of $10^{-3}$ to decrease the sigmoid cross-entropy loss. Except the learning rate, we employed the same parameter values given in~\cite{simonyan2014very, he2016deep, Sumbul:2019}. The batch size is set to $256$ for ResNet152 and to $500$ for all other models used in the experiments. We applied training from scratch for $100$ epochs, while the final layers of the pre-trained models were fine-tuned separately on each modality for $10$ epochs. 
For all the models, we added a fully connected layer that includes $19$ neurons at the end of the network for the classification. For image retrieval, we extracted image features from the considered models and applied similarity matching of the features based on the $\chi^2$-distance measure. We performed various experiments to analyze the effectiveness of: i) learning from BigEarthNet-MM directly (through training from scratch) instead of using the pre-trained models on ImageNet; and ii) state-of-the-art CNN models trained and evaluated on BigEarthNet-MM. To use the pre-trained models on ImageNet, we used the late fusion of separately fine-tuned models on Sentinel-1 and Sentinel-2 patches. In the experiments, we did not use the Sentinel-2 patches that are fully covered by seasonal snow, cloud, and cloud shadow. After the arrangements of the new class nomenclature, $57$ pairs among the $590,326$ pairs are not associated with any LCLU labels. these pairs are not used in the experiments. We divided the remaining dataset into: i) the training set of $269,695$ pairs of patches, ii) validation set of $ 123,723 $ pairs of patches, and iii) the test set of $ 125,866 $ pairs of patches.

We performed our experiments on a cluster of 4 NVIDIA Tesla V100 GPUs. The results of multi-modal multi-label image classification were provided in terms of four performance metrics: 1) Hamming loss ($HL$); 2) one-error ($OE$); 3) recall ($R$); and 4) $F_2$-Score ($F_2$). For a detailed description of the considered metrics, the reader is referred to \cite{Sumbul:2019}.
\begin{table}[t]
\renewcommand{\arraystretch}{0.1}
\setlength{\tabcolsep}{3.5pt}
\centering
\footnotesize
\caption{Class-based $F_2$ Scores ($\%$) obtained when: i) transfer learning from ImageNet and ii) direct learning from BigEarthNet-MM are used for multi-modal multi-label image classification.}
\label{table:comp_pretrained_class1}
\begin{tabular}{@{}lccc}
\toprule
\textbf{Class} & \parbox{2.3cm}{\centering \textbf{Transfer Learning\\From ImageNet}} & \parbox{2.2cm}{\centering \textbf{Learning From BigEarthNet-MM}} \\
& & \\\toprule
Urban fabric & \percentage{0.5627467400406747} &\percentageBest{0.7198579529182656}\\ \midrule
\parbox{3.51cm}{\raggedright Industrial or commercial units}  &\percentage{0.30980647259717026} & \percentageBest{0.43211745118813377} \tabularnewline \midrule
Arable land & \percentage{0.8004703557646026} &  \percentageBest{0.8361905573183976} \tabularnewline \midrule
Permanent crops & \percentage{0.043189489347802416} &  \percentageBest{0.5551538160233812} \tabularnewline \midrule
Pastures & \percentage{0.5097641505143574} &  \percentageBest{0.747701017770069} \tabularnewline \midrule
\parbox{3.51cm}{\raggedright Complex cultivation patterns} &  \percentage{0.3629244653580082} & \percentageBest{0.6202635980250154} \tabularnewline \midrule
\parbox{3.51cm}{\raggedright Land principally occupied by agriculture, with significant areas of natural vegetation}  &\percentage{0.3036166468762071} & \percentageBest{0.6062909967743372} \tabularnewline  \midrule
Agro-forestry areas & \percentage{0.021255442421749118} &  \percentageBest{0.7187025690033166} \tabularnewline \midrule
Broad-leaved forest & \percentage{0.4283262589739559} &  \percentageBest{0.7538791670797059} \tabularnewline \midrule
Coniferous forest & \percentage{0.7547233206688894} &  \percentageBest{0.8632052926823133} \tabularnewline \midrule
Mixed forest & \percentage{0.7218610041177281} &  \percentageBest{0.8130655648560845} \tabularnewline \midrule
\parbox{3.51cm}{\raggedright Natural grassland and sparsely vegetated areas} & \percentage{0.14106313435403406} &  \percentageBest{0.4388195407612883} \tabularnewline \midrule
\parbox{3.51cm}{\raggedright Moors, heathland and sclerophyllous vegetation} &  \percentage{0.05293367346938776} & \percentageBest{0.5991496955700188} \tabularnewline \midrule
\parbox{3.51cm}{\raggedright Transitional woodland-shrub} &  \percentage{0.412335078028896} & \percentageBest{0.6420896955516436} \tabularnewline \midrule
\parbox{3.51cm}{\raggedright Beaches, dunes, sands} &  \percentage{0.4367469879518072} & \percentageBest{0.6338652482269503} \tabularnewline \midrule
Inland wetlands & \percentage{0.08201979472140762} & \percentageBest{0.5780705153041457} \tabularnewline \midrule
Coastal wetlands & \percentage{0.04792332268370607} &  \percentageBest{0.42233357193987114} \tabularnewline \midrule
Inland waters & \percentage{0.6322630171797655} &  \percentageBest{0.8210417870693771} \tabularnewline \midrule
Marine waters & \percentage{0.9399161206349913} &  \percentageBest{0.9719544943006452} \tabularnewline \midrule
\midrule
\textit{Average} & \percentage{0.39809923556342847} &  \percentageBest{0.6723027648612083} \tabularnewline \midrule
\bottomrule
\end{tabular}
\end{table}
\subsection{Experimental Results}

\subsubsection{Comparison among the Strategies of Learning directly from BigEarthNet-MM and Transfer Learning from the ImageNet} 
In the first set of experiments, we compare the effectiveness of learning directly from BigEarthNet-MM with respect to transfer learning from ImageNet. To this end, transfer learning strategy is applied by using the pre-trained ResNet50 model trained on ImageNet, while direct learning strategy is employed by using the ResNet50 trained from scratch on BigEarthNet-MM. Table~\ref{table:comp_pretrained_class1} shows the class-based $F_2$ classification scores (known also as macro-averaged $F_2$ scores \cite{Sumbul:2019}). By analyzing the table, one can see that learning directly from BigEarthNet-MM achieves the highest score for each class compared to the transfer learning strategy. As an example, learning directly from BigEarthNet-MM provides more than $12\%$ and $25\%$ higher scores for the classes \textit{Industrial or commercial units} and \textit{Complex cultivation patterns}, respectively, compared to the transfer learning strategy. The difference in performance between these learning strategies is more evident for more complex LULC classes. As an example, learning directly from BigEarthNet-MM improves the $F_2$ scores more than $54\%$ and $69\%$ for the classes \textit{Moors, heathland and sclerophyllous vegetation} and \textit{Agro-forestry areas}, respectively. 

\begin{figure*}[t]
  \centering
  \includegraphics[width=\linewidth]{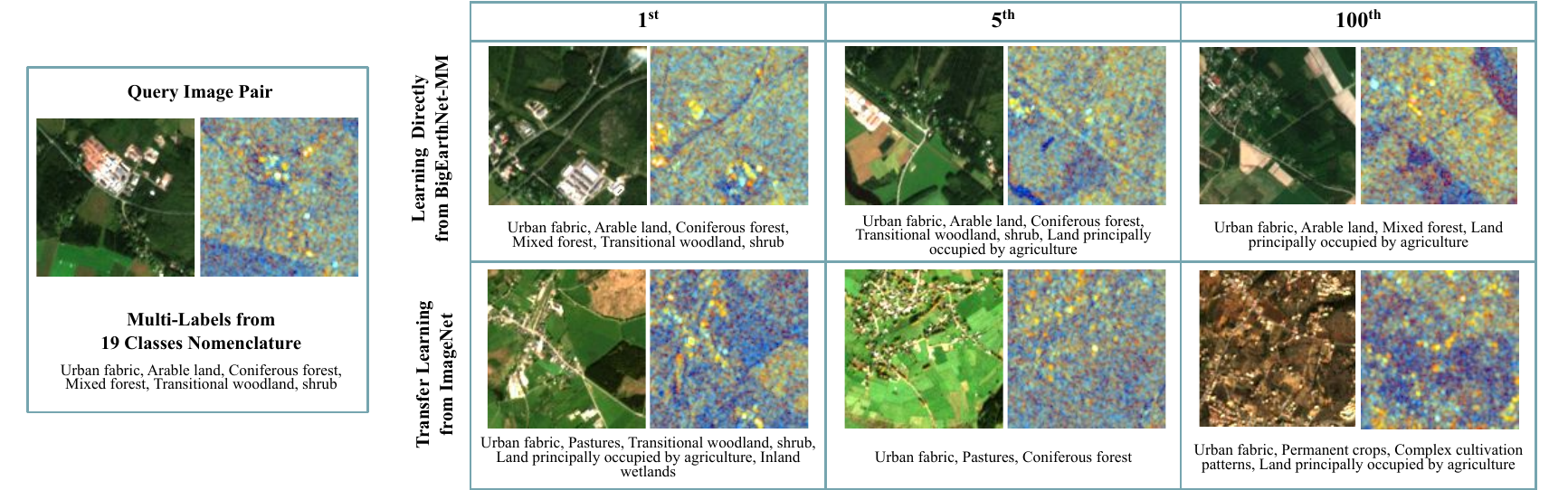}
  \caption{An example of a query pair from the BigEarthNet-MM archive and retrieved image pairs obtained by using: 1) direct learning from BigEarthNet-MM; and 2) transfer learning from ImageNet in the framework of content-based multi-modal multi-label image retrieval.}
  \label{fig:retrieval}
\end{figure*}

In the content of image retrieval, Fig.~\ref{fig:retrieval} shows an example of a query pair and the retrieved pairs of images by these strategies. By assessing the figure, one can observe that when learning is achieved directly from BigEarthNet-MM, the semantically more similar pairs of images are retrieved, containing the \textit{Urban fabric} and \textit{Arable land} classes present in the query. Learning directly from BigEarthNet-MM leads to retrieval of a similar pair to the query even at the 100\textsuperscript{th} retrieval order. However, using transfer learning strategy results in retrieval of pairs that contain \textit{Urban fabric} and \textit{Arable land} classes which are not present in the query pair. One can observe this behavior even at the 5\textsuperscript{th} retrieved pair.

The main reasons of the success of directly learning from BigEarthNet-MM are due to the fact that: 1) transfer learning from ImageNet limits the accurate characterization of the spectral content of RS images; 2) fine-tuning the pre-trained model on ImageNet by using RS images can not be sufficient to eliminate the semantic gap since the category labels present in ImageNet are different from the land-cover class labels present in BigEarthNet-MM; and 3) the pre-trained model was trained for a single-label image classification scenario, and thus limits the accurate characterization of the multiple land cover classes present in BigEarthNet-MM. 

\begin{table}[t]
\renewcommand{\arraystretch}{1}
\setlength{\tabcolsep}{12pt}
\centering
\footnotesize
\caption{Overall Multi-Label Classification Results Under Different Metrics and DL Models for BigEarthNet-MM.}
\label{table:comp_methods}
\begin{tabular}{@{}lcccc@{}}
\toprule
Model &$HL$&$OE$($\%$)&$R$($\%$)&$F_2$($\%$) \tabularnewline
\toprule
 VGG16 & \round{0.07778614} & \percentage{0.073522635} & \percentage{0.76969254} & \percentage{0.76182824} \\
 VGG19 & \round{0.07973057} & \percentage{0.081205405} & \percentage{0.7617114} & \percentage{0.75348103} \\
 ResNet50  & \round{0.07409968} & \percentageBest{0.05930116} & \percentageBest{0.8004607} & \percentageBest{0.78731185} \\
 ResNet101  & \round{0.07393871} & \percentage{0.06461634} & \percentage{0.7884641} & \percentage{0.778826} \\
 ResNet152  & \roundBest{0.07277455} & \percentage{0.06419525} & \percentage{0.78128016} & \percentage{0.7745805} \\
 \midrule
\bottomrule
\end{tabular}
\end{table}
\subsubsection{Comparison of State-of-the-Art CNN Models}
In the second set of experiments, we compare the effectiveness of the VGG and the ResNet models in the framework of multi-modal multi-label classification. Table~\ref{table:comp_methods} shows the overall classification results under different metrics (which are the sample-averaged scores \cite{Sumbul:2019}). By analyzing the table, one can observe that the ResNet model provides the highest scores in all metrics. As an example, ResNet50 achieves more than $2\%$ higher recall and $F_2$ scores compared to VGG models. This improvement is due to the residual connections of the ResNet model and their increased depth in terms of the number of layers compared to the VGG model. Increasing the depth of the considering models does not significantly affect the performances, i.e., similar scores are obtained in all the metrics under different depth values of the same model.

\section{Discussion and Conclusion}
\label{sec:conclusion}
In this paper, we have presented the BigEarthNet-MM benchmark archive that contains 590,326 pairs of Sentinel-1 and Sentinel-2 image patches with a new CLC-based class-nomenclature to pay more justice to the properties of the considered images. BigEarthNet-MM makes a significant advancement for the use of DL in RS, opening up promising directions to support research studies in the framework of multi-modal multi-label RS image scene classification and retrieval. BigEarthNet-MM is suitable to assess DL based methods for: i) learning from class-imbalanced multi-modal data (since the LCLU classes are not equally represented in BigEarthNet-MM); ii) transfer learning (since BigEarthNet-MM currently contains only pairs of images from a small number of European countries); and iii) also on unsupervised, self-supervised and semi-supervised multi-modal learning for information discovery from big data archives. 

It is worth noting that BigEarthNet-MM has limitations for the RS applications that require time-series data to accurately describe LCLU classes, such as \textit{Non-irrigated arable land}, \textit{Permanently irrigated land}. We would like to also note that some Sentinel-1 image patches can be contaminated by artefacts caused by either well-known Radio-Frequency-Interference \cite{tao2019mitigation} or other dataset related issues, which are independent from the pre-processing steps applied while constructing BigEarthNet-MM. As a final remark, we would like to point out that due to the use of labels from the CLC map, the BigEarthNet-MM archive can be extended to a larger scale within Europe with zero-annotation cost. As a future development of this work, we plan to enrich the BigEarthNet-MM archive by extending it to whole Europe. 

\section*{Acknowledgment}
This work is funded by the European Research Council (ERC) through the ERC-2017-STG BigEarth Project under Grant 759764 and by the German Ministry for Education and Research as BIFOLD - Berlin Institute for the Foundations of Learning and Data (01IS18025A). The authors from DGT are supported through FCT (Fundação para a Ciência e a Tecnologia) - UIDB/04152/2020 - Centro de Investigação em Gestão de Informação (MagIC).
\bibliographystyle{IEEEtran}
\bibliography{bib_definitions,bibliography}
\end{document}